\documentclass[10pt,twocolumn,letterpaper]{article}

\usepackage{wacv}              

\usepackage{graphicx}
\usepackage{amsmath}
\usepackage{amssymb}
\usepackage{booktabs}
\usepackage{comment}
\usepackage[accsupp]{axessibility}

\usepackage[pagebackref,breaklinks,colorlinks]{hyperref}

\usepackage[capitalize]{cleveref}
\crefname{section}{Sec.}{Secs.}
\Crefname{section}{Section}{Sections}
\Crefname{table}{Table}{Tables}
\crefname{table}{Tab.}{Tabs.}

\begin{document}

\title{Real Time GAZED: Online Shot Selection and Editing of Virtual Cameras \\ from Wide-Angle Monocular Video Recordings}

\author{Sudheer Achary, Rohit Girmaji, Adhiraj Anil Deshmukh, Vineet Gandhi\\
CVIT, KCIS, IIIT Hyderabad, India\\
{\tt\small sudheer.achary@research.iiit.ac.in, rohit.girmaji@research.iiit.ac.in}
 }
\maketitle

\begin{abstract}
Eliminating time-consuming post-production processes and delivering high-quality videos in today's fast-paced digital landscape are the key advantages of real-time approaches. To address these needs, we present Real Time GAZED: a real-time adaptation of the GAZED framework integrated with CineFilter, a novel real-time camera trajectory stabilization approach. It enables users to create professionally edited videos in real-time. Comparative evaluations against baseline methods, including the non-real-time GAZED, demonstrate that Real Time GAZED achieves similar editing results, ensuring high-quality video output. Furthermore, a user study confirms the aesthetic quality of the video edits produced by the Real Time GAZED approach. With these advancements in real-time camera trajectory optimization and video editing presented, the demand for immediate and dynamic content creation in industries such as live broadcasting, sports coverage, news reporting, and social media content creation can be met more efficiently.
\end{abstract}

\section{Introduction}
Creating professional recordings of live stage performances involves skilled camera operators who capture the performance from various angles. These camera feeds are edited to create a polished and engaging final product. However, generating these professional edits is a challenging task. Firstly, operating cameras during a live performance is difficult even for experts, as there are no second chances to retake footage, and there are limitations on camera angles due to the impracticality of using large equipment like trolleys or cranes. Secondly, manual video editing is a slow and laborious process that requires the expertise of skilled editors. Producing professional recordings of live performances requires a professional camera crew, multiple cameras and equipment, and experienced editors, which increases the complexity and costs of the process.

To this end, small-scale productions end up using a fixed wide-angle camera placed at a sufficiently large distance to capture the entire stage. These static recordings are suitable for archiving purposes. While they provide an overall understanding of the context, the distant camera feed fails to showcase essential elements of cinematic storytelling, such as close-up shots of faces, characters' emotions and actions, and important actor interactions for creating an engaging experience.
\begingroup
\setlength{\tabcolsep}{1pt}
\begin{figure}
    \centering
    \begin{tabular}{c c c}
        \includegraphics[width=0.3\linewidth]{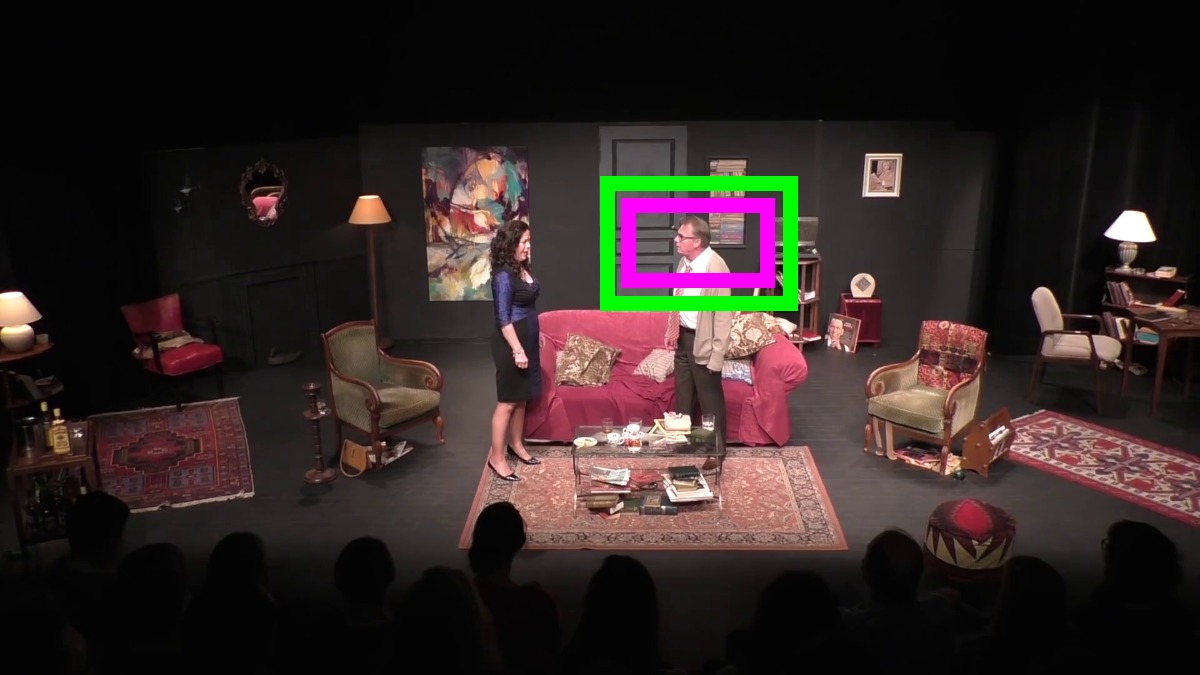} &
        \includegraphics[width=0.3\linewidth]{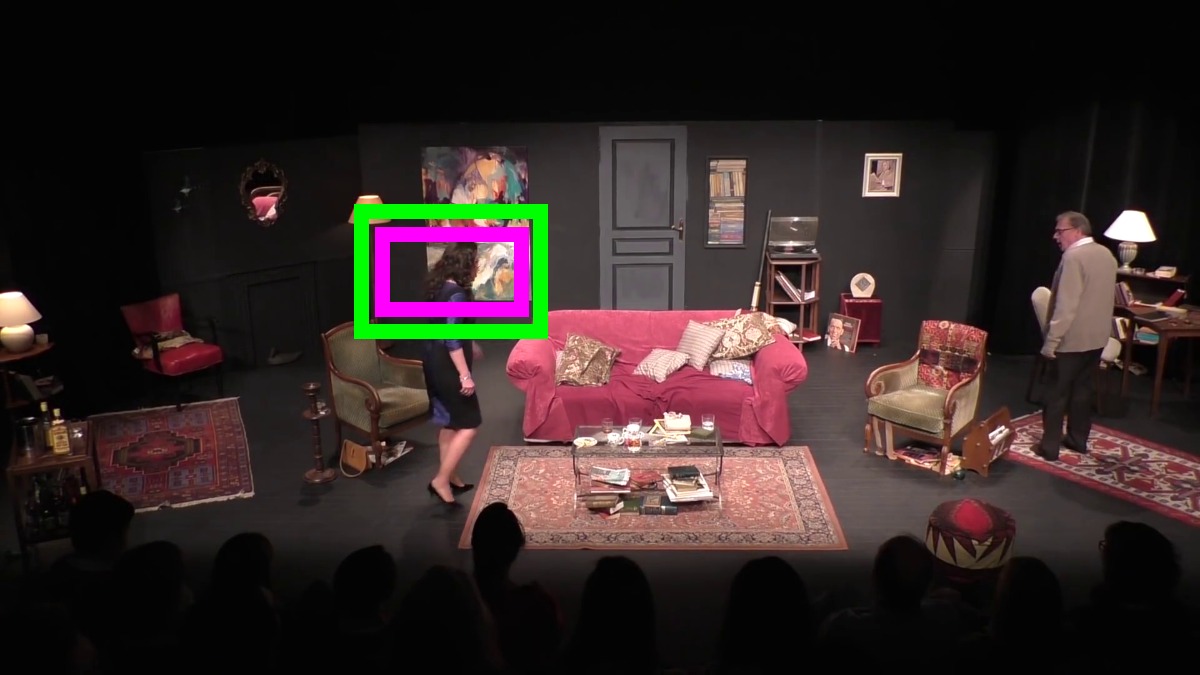} &
        \includegraphics[width=0.3\linewidth]{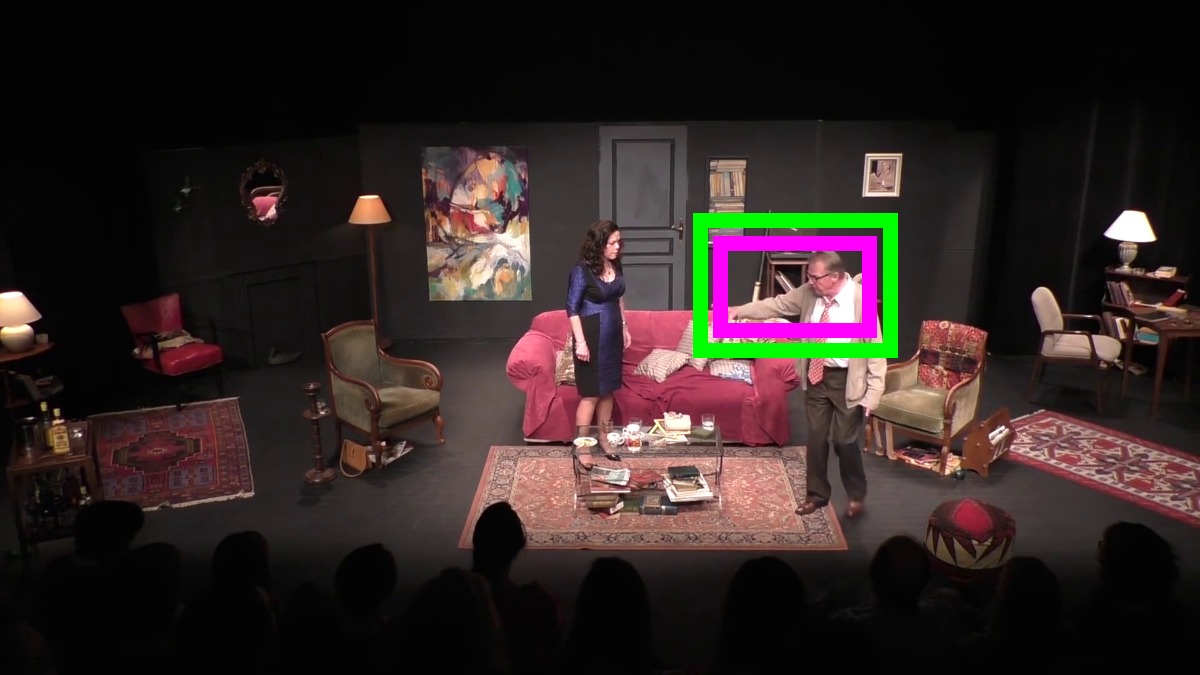}
    \end{tabular}

    \centering
    \begin{tabular}{c}
        \includegraphics[width=0.95\linewidth]{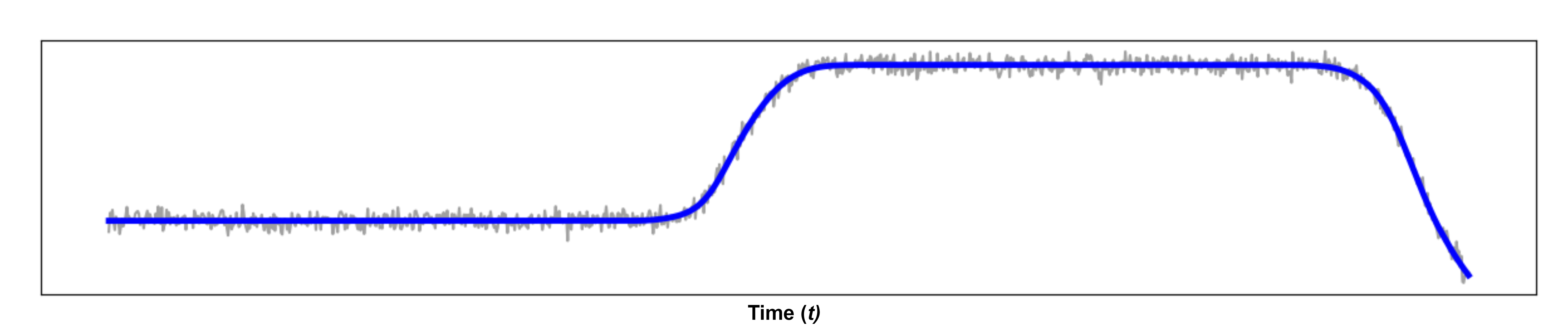}
    \end{tabular}

    \centering
    \begin{tabular}{c c c}
        \includegraphics[width=0.3\linewidth]{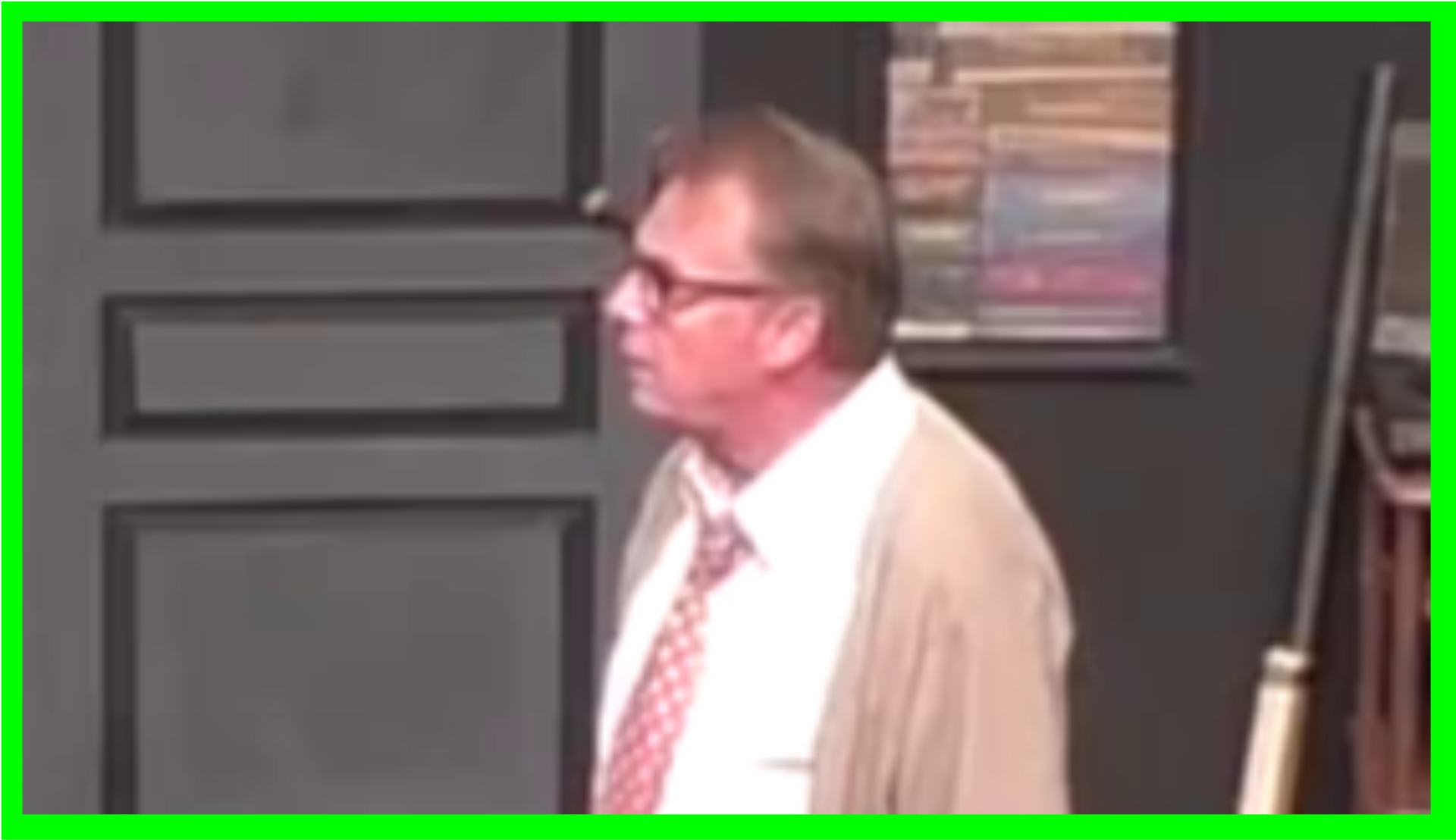} &
        \includegraphics[width=0.3\linewidth]{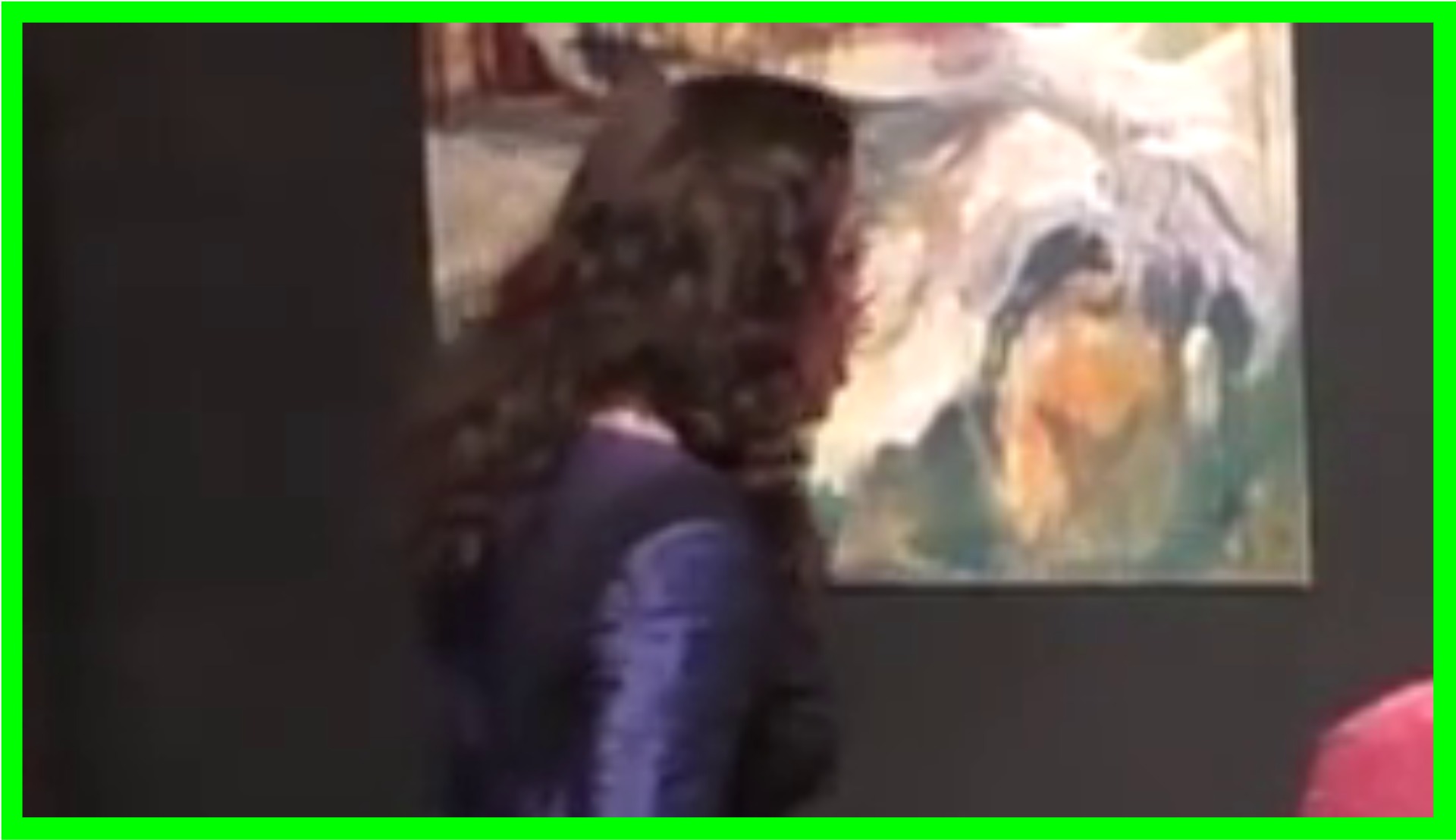} &
        \includegraphics[width=0.3\linewidth]{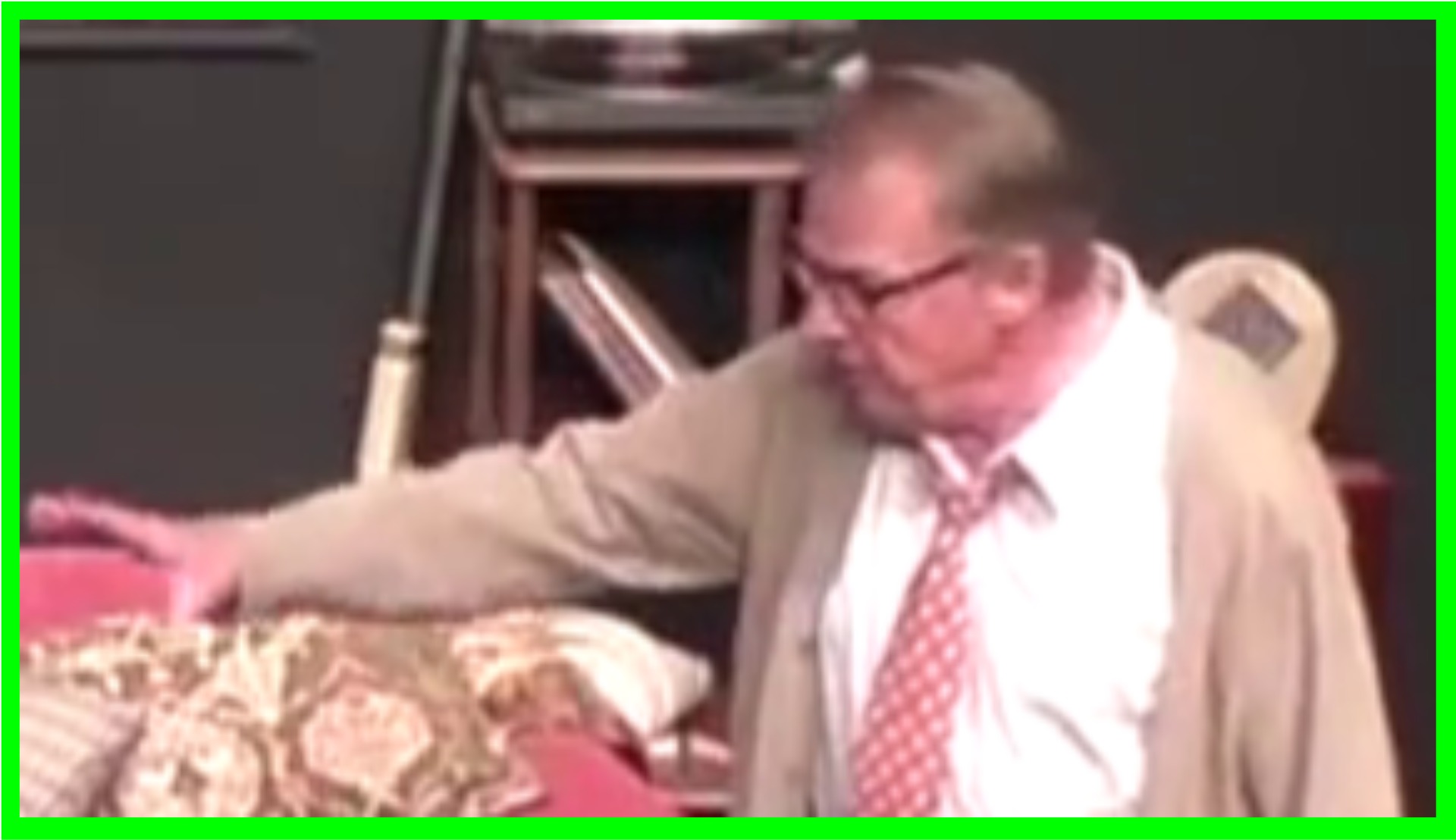}
    \end{tabular}
    \caption{The initial row of bounding boxes illustrates the shots carefully chosen by the Real Time GAZED system, indicated by the green boxes, and the GAZED system, represented by the pink boxes. The second row depicts the noisy (in gray) and real-time stabilized (in blue) actor's trajectory along horizontal (X) direction in the video. The third row showcases the cropped shots from frames.}
    \label{fig:Figure1}
\end{figure}
\endgroup

A recent effort called GAZED~\cite{Gazed20} demonstrated promise in automated editing from a single static high-resolution recording employing the user's gaze. They first simulate multiple virtual cameras by moving a cropping window inside the original static recording\cite{gandhi2014multi} and then perform camera selection among the simulated camera feeds. Gaze information helps identify essential areas within the scene, which are assigned gaze potentials that quantify the significance of the available shots. These gaze potentials are combined with other factors that adhere to cinematic principles, such as avoiding abrupt cuts, maintaining rhythm, avoiding transient shots, etc. Dynamic programming is used to solve the discrete optimization problem of camera selection.

The major limitation of GAZED is that both the virtual camera simulation and camera selection procedures are posed as an offline optimization, limiting its usage in live streaming applications (like music concerts, staged dance performances, etc.). In this work, we propose Real Time GAZED, which extends the GAZED framework to such scenarios by remodeling the optimization algorithms into real-time setup, with a minimal look ahead into the future. For virtual camera simulation, we adapt Cinefilter~\cite{CineFilter} and propose a novel formulation for real-time camera selection, with additional constraints enforcing continuity across time. The Fig. \ref{fig:Figure1} concisely represents the core functionalities encompassed by the Real Time GAZED and GAZED video editing pipelines. 

A user study involving 8 participants was conducted to evaluate the system's real-time performance. Multiple edited versions of stage performance recordings are edited using Real Time GAZED and compared against several baseline methods. The user study results demonstrate that Real Time GAZED outperforms the baseline methods in terms of editing quality and performs equally well compared to GAZED. Thus, even with a real-time version, incorporating gaze information and other cinematic principles leads to more effective and engaging video edits. Our main contributions can be summarized as follows:

\begin{enumerate}
    \item We created an end-to-end cinematic editing pipeline that operates in real-time. It allows for generating professional quality videos from a static camera recording. The approach involves selecting shots based on an objective function that incorporates gaze potentials and adheres to cinematic principles and shot continuity constraints. This system empowers even novice users to create polished and well-edited videos using their eye gaze data and an affordable desktop eye tracker.
    
    \item We conducted a comprehensive user study to validate the method's effectiveness compared to various editing baselines. The results demonstrate that users prefer the outputs generated regarding several attributes that characterize the editing quality.
\end{enumerate}

\section{Related Work}
Many scholarly articles have taken on the challenge of video editing by treating it as a discrete optimization problem \cite{IntelligentVirtualCinematography} \cite{ContinuityEditing} \cite{ComputationalModelFilmEditing} \cite{VirtualDirector}. Most of these studies have embraced dynamic programming techniques to determine the optimal amalgamation of shots that enhance viewer engagement. However, it is worth noting that these approaches enjoy the luxury of unrestricted camera placement and movement within their 3D environments - an advantage that is absent from our methodology. Moreover, it is impractical to implement these approaches in real-time scenarios where retakes are not feasible.


Meratbi et al. \cite{ComputationalModelFilmEditing} employs a Hidden Markov Model for editing using established film shot transition probabilities, focusing on dialogue scenes with manual movie annotation. In contrast, our approach offers versatile editing without scene-specific limits, embracing creative potential. Galvane et al. \cite{ContinuityEditing} contribute significantly, addressing cut placement, rhythm, and continuity. We draw from these innovations, but stage performance challenges are distinct, lacking 3D scene data and camera freedom.


Ranjan et al. \cite{ImprovingMeetingCapture} transformed group meeting editing using cues like speaker detection, posture, and head orientation to establish effective editing rules. For instance, switching to close-ups when speakers change or using overview shots during conversations. Doubek et al. \cite{CinematographicRules} focused on camera selection for surveillance. These studies revolutionize video editing and camera choices, enhancing engagement. Sports event camera selection is also advanced \cite{ComputationalSportsBroadcasting} \cite{PersonalizedProductionOfBasketball} \cite{CameraSelectionForBroadcasting} \cite{AutomaticComposition}, often employing Hidden Markov Models \cite{PersonalizedProductionOfBasketball} \cite{AutomaticComposition} for diverse viewpoints.

Many studies explore learning-based video editing methods. For instance, Anyi et al. \cite{TransformerEditing} excel in camera selection for broadcasting but miss actor emotions captured by traditional editing. Hui-Yin Wu et al. \cite{AutomaticEditing} identify optimal meeting shots via annotated videos, with high costs. In contrast, our approach offers efficient, real-time deployment with low computational needs and memory usage. Studies like \cite{GazeDrivenVideoEditing} \cite{WatchToEdit} use eye gaze data for video retargeting, adapting content between display devices. Current methods focus on horizontal adjustments, leading to poorly composed frames and odd zoom effects. GAZED \cite{Gazed20} offers improved shot selection using gaze, but it's limited to offline editing. Our approach enhances GAZED for real-time use, addressing this limitation effectively.

\section{Method}
In Real Time GAZED, we enhanced the GAZED video editing pipeline by adjusting its shot generation and shot selection components. Firstly, we have replaced the offline actor trajectory stabilization with a real-time actor trajectory stabilization \cite{CineFilter} for the shot generation component. In the shot selection component, we have introduced a shot continuity constraint in addition to the cost matrix and previously selected shots. This constraint ensures the selected shots flow smoothly and maintain a coherent sequence. By incorporating these modifications, we made it possible to use these components in real-time settings. They rely on contextual information available at any given moment and rather not use information across time. Below, we briefly explain the functionality of these components.

\subsection{GAZED}
The video editing pipeline of GAZED \cite{Gazed20} comprises two components: shot generation and shot selection. The shot generation component produces various combinations of shots featuring the actor, utilizing de-noised data obtained from a widely recognized human tracker called ByteTrack \cite{zhang2022bytetrack}. An offline trajectory stabilizer is employed to refine the actor's trajectories to enhance the composition of shots. The shot selection stage treats it as a convex optimization problem. It employs dynamic programming techniques to identify and select the most suitable shot at a given time $t$. The approach uses an offline strategy that involves backtracking through the entire video to identify a shot from the optimal shot cost path. 

\subsection{Shot Generation}
The shot generation component in the GAZED video editing pipeline utilizes a wide-angle recording captured by a stationary camera, providing a complete view of the entire scene. Each frame in this input video serves as a master shot. Through the application of a virtual camera simulation technique \cite{gandhi2014multi} known as multi-virtual pan-tilt-zoom (PTZ) cameras, possible shots are automatically generated for actors within each frame. It involves using virtual PTZ cameras that focus on specific actors or groups of actors. These cameras create zoomed-in shots that bring a sense of depth and intimacy to the original wide-angle recording. To determine the actor positions within each master shot, we leverage the information provided by bounding boxes obtained through the use of a human tracker \cite{zhang2022bytetrack}. 

\subsubsection{CineFilter}
However, these bounding boxes obtained from the tracker may contain noise and exhibit jitter. Instead of relying on an offline trajectory stabilization approaches \cite{grundmann2011auto,gandhi2014multi,tang2019joint}, we employ \emph{CineFilter} \cite{CineFilter} a novel approach for real-time camera trajectory optimization. It comprises two online filtering methods: \emph{CineConvex} and \emph{CineCNN}. CineConvex utilizes a sliding window-based convex optimization formulation, while CineCNN employs a convolutional neural network as an encoder-decoder model. Both methods are motivated by cinematographic principles, producing smooth and natural camera trajectories. With a minor latency of half a second, CineConvex operates at approximately 250 frames per second (fps), while CineCNN achieves an impressive speed of 1000 fps, making them highly suitable for real-time applications. It eliminates noise, jitter, and residual motion while emulating an ideal camera trajectory composed of three segments: static segments, constant velocity segments, and segments with constant acceleration, resulting in a smooth trajectory. The middle row of Fig. \ref{fig:Figure1} portrays an noisy and the corresponding real-time stabilized trajectory.

In generating shots with real-time actor trajectory stabilization, we employ the CineCNN model as our preferred choice. This model stands out due to its learning-based, lightweight nature and fast processing capabilities.

With real-time actor trajectory stabilization integration, the shot generation component can operate in real-time without relying on temporal video information. Instead, it focuses solely on the actors present within each frame at a specific moment, stabilizes their trajectories, and rapidly generates shots without considering the context across multiple frames. This approach instantaneously captures the scene's essence, enabling dynamic and on-the-fly shot selection. Consequently, the process is fast and efficient. When processing an input video, we generate a comprehensive set of possible shots for every combination of performers in the scene. For a video with $n$ performers, we create $n*(n+1)/2$ combinations of shots. We generate $n$ number of 1-shots for sequences with N actors, followed by $N-1$ number of 2-shots, $N-2$ number of 3-shot type, and so on, capturing different arrangements of performers. We choose shots featuring neighboring actors, resulting in $N-2$ instances of 2-shots. This selection employs a sliding window approach, avoiding the need for 2 permutations of N actors. This process applies similarly to 3-shot sequences and beyond.

\begin{figure}
    \centering
    \includegraphics[width=0.75\linewidth]{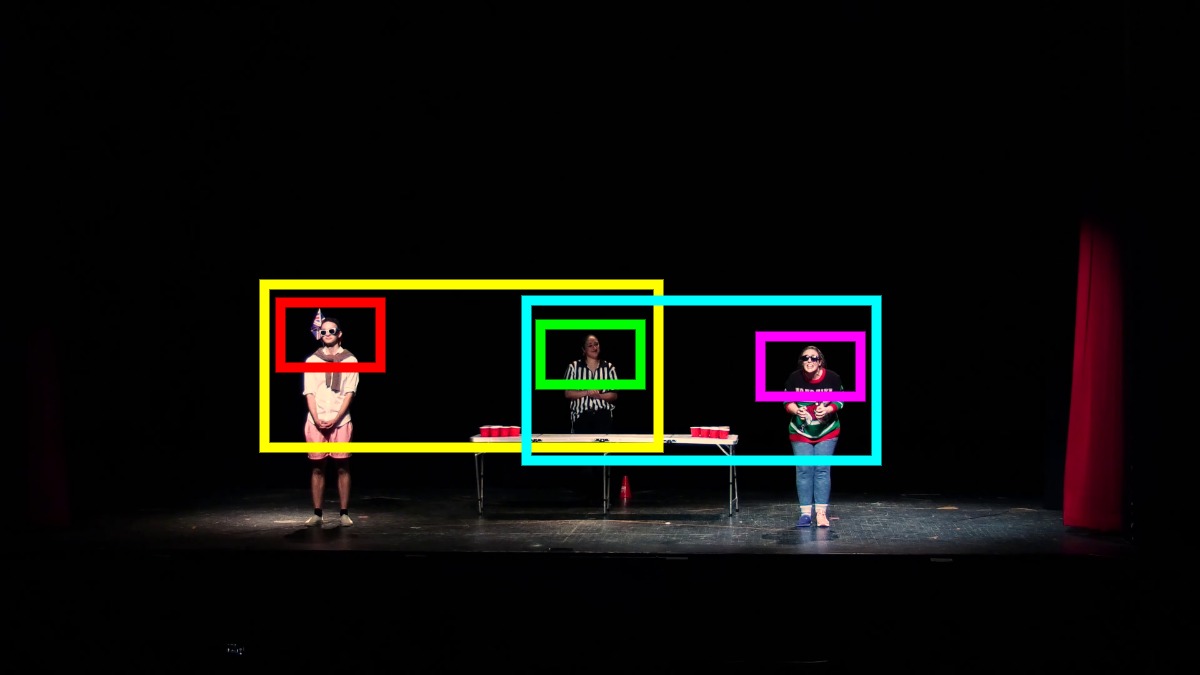}
    \caption{The figure illustrates the various bounding boxes generated within a frame. These bounding boxes serve as virtually simulated cameras, capturing different perspectives and compositions from a single frame. These generated shots are used in Real Time GAZED algorithm.}
    \label{fig:Figure3}
\end{figure}

The Fig. \ref{fig:Figure3} portrays the possible shot combinations within a single frame featuring three actors. We utilize a Medium Shot (MS) for single actor shots (1-shots) to ensure the generated shots are visually appealing. A medium shot frames the performer from head to waist, while a medium closeup focuses from head to mid-chest, offering an intimate perspective. For sequences involving multiple actors, we employ a Full Shot (FS) that captures each performer from head to toe, providing a comprehensive view of the group's dynamics. By implementing these techniques, we enhance the visual impact of the edited videos and create a more immersive viewing experience for the audience. We denote a set of shots generated from a frame (master-shot) by $(S)$.
\begin{equation}
    S = \{s_{i}\}_{i=1}^{n*(n+1)/2}
    \label{eqn:Equation1}
\end{equation}

To overcome the low resolution of shots generated from the original video, Its recommended to capture video in a higher resolution. Modern high-performance CPUs and GPUs smoothly handle 4K or 8K video processing. Dedicated graphics cards and specialized video chips accelerate tasks like decoding, encoding, and editing. Quick access to large video files requires fast SSDs with high read and write speeds.

\subsection{Shot Selection}
In our video editing pipeline, the next crucial step after generating shots is selecting the most compelling shot that effectively tells the story at each moment. However, we couldn't directly utilize the shot selection component from the GAZED pipeline as it heavily relies on the complete temporal information of the video. To overcome this limitation, we modified the shot selection component by considering only a small time frame, typically ranging from half to one second, instead of the entire video. Additionally, we introduced an extra penalty term called shot continuity to ensure a smooth transition between shots.

The shot selection process is treated as a discrete optimization problem, where we assess the importance of each of the multiple shots generated for every video frame. During this assessment, we adhere to fundamental cinematic principles such as avoiding abrupt cuts between overlapping shots (jump cuts), preventing rapid shot transitions, and maintaining a cohesive cutting rhythm. To determine the importance of each shot at a given moment, we rely on eye gaze data collected using an eye-tracking device. Moreover, we incorporate cinematic principles into the optimization process through penalty terms that guide the shot selection. By making these modifications and incorporating eye gaze data and cinematic principles, we enhance the overall editing process, ensuring that the selected shots effectively convey the story and captivate the audience.

\begin{figure}[t]
    \begin{tabular}{c c}
        \includegraphics[width=0.6\linewidth]{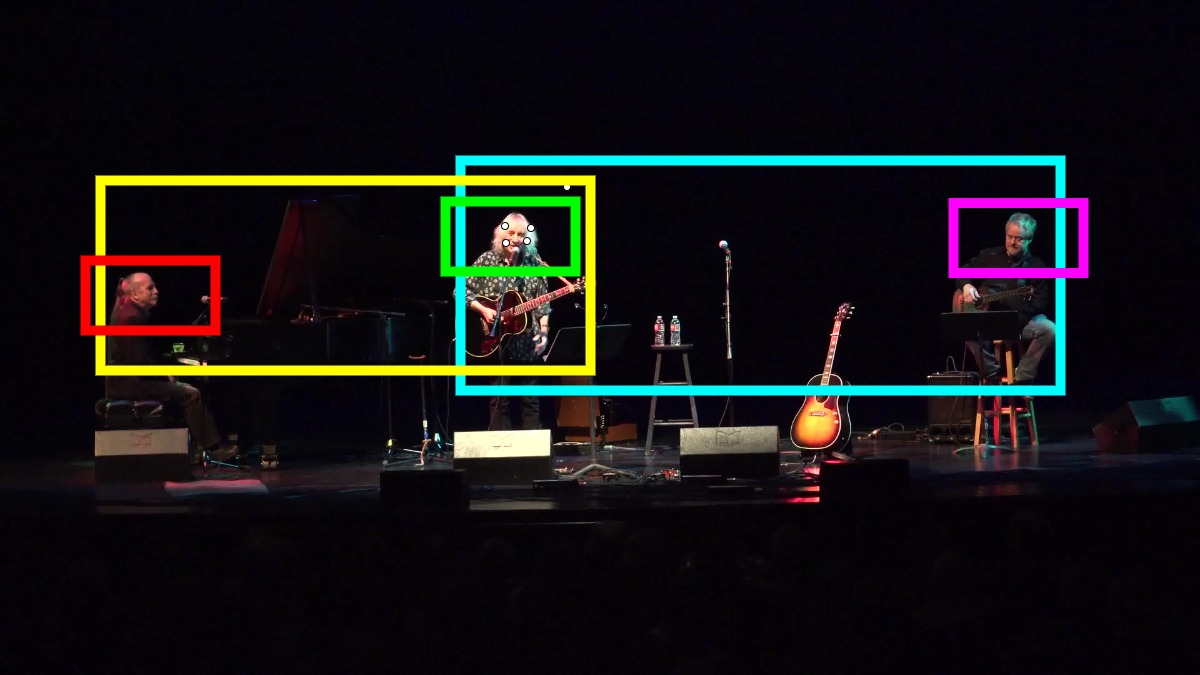} &
        \includegraphics[width=0.3\linewidth]{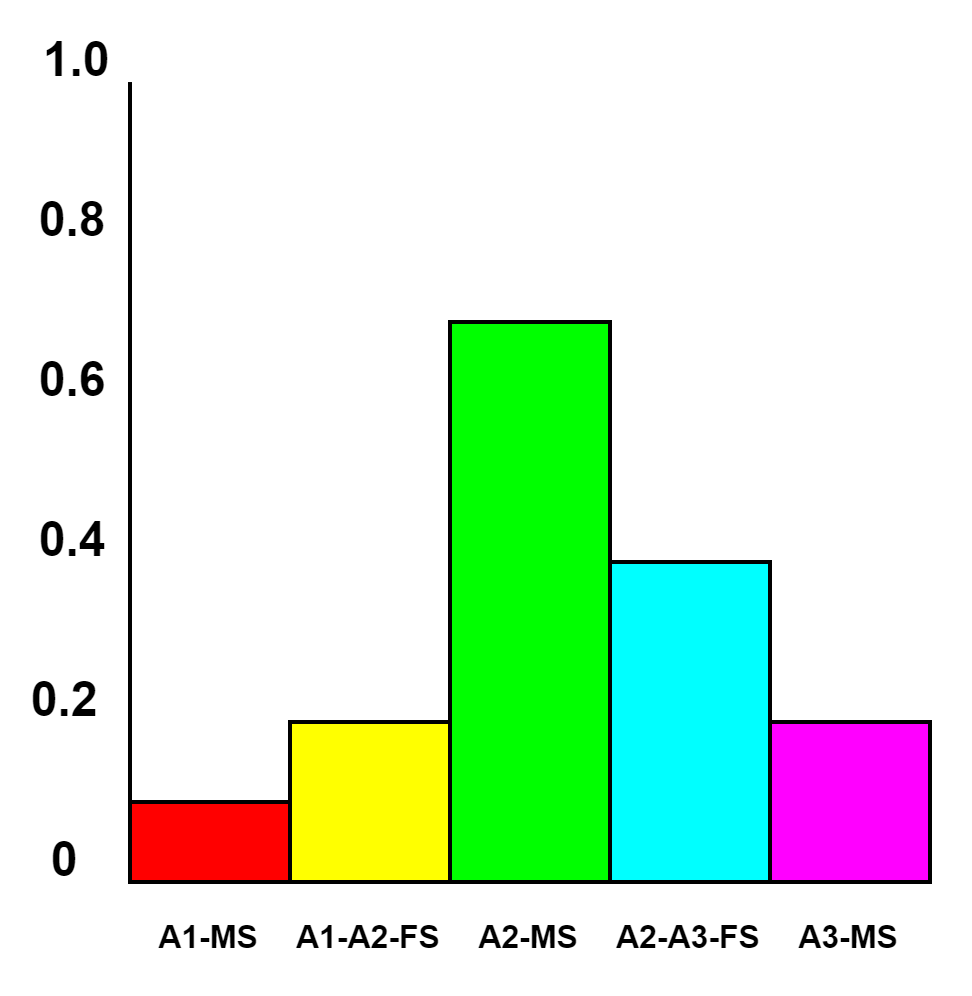}
    \end{tabular}    
    \caption{The figure showcases the behavior of the gaze potential function in response to human gaze. It provides a visual representation of this interaction by highlighting white dots that indicate the precise locations where the human gaze is directed within the frame. To better understand the influence of gaze on the scene, accompanying histograms are displayed beside each frame. These histograms present the gaze potential cost associated with each bounding box in the scene. To make it even more intuitive, the color-coded bars in the histograms correspond to the respective bounding boxes, allowing for a quick and easy comparison. For instance, the green bar in the histogram represents the gaze potential cost of the bounding box highlighted in green.}
    \label{fig:Figure4}
\end{figure}

For a scene with $n$ actors, the editing graph consists of $n*(n+1)/2$ nodes at each frame $t$, where each node represents a shot and edges across time steps represent a transition from one shot to another (denoting a cut) or to itself (no cut). Formally, given a sequence of frames $t = [1..T]$ the set of generated shots  $S^{t} = \{ s_{i}^{t}\}_{i=1}^{n*(n+1)/2}$ and the raw gaze data $g_{k}^{t}$ corresponding to user $k$ at time $t$, our algorithm selects a sequence of shots $\epsilon = \{s^{t}\}$ where $s^{t} \in S^{t}$, by minimizing the following objective function:

\begin{equation}
    E(\epsilon) = \sum_{t=1}^{T} -ln(G(s^{t})) + \sum_{t=2}^{T} E_{e}(s^{t-1}, s^{t})
    \label{eqn:Equation2}
\end{equation}

where $E_{e}(s^{t-1}, s^{t})$ denotes cost for switching from one shot to another and $G(s^{t})$ is a unary cost that represents the gaze potential (modeling importance) for each shot.

\subsubsection{Gaze Potential}
Each generated shot is assigned a score that helps the optimization algorithm find the most optimal path through the editing graph. When editing a video, ensuring that the final result captures each scene's original narrative is essential. Unlike previous methods \cite{ComputationalVideoEditing}, \cite{ContinuityEditing} that rely on additional metadata or computational features to estimate actions or emotions in a shot, which often overlook high-level scene semantics that humans are sensitive to, we utilize gaze data recorded from users. This approach has proven to be effective in accurately localizing focal scene events. We choose to use the Gaze Potential component from GAZED as it has several advantages. It is fast, does not depend on the temporal context within the video, and can be computed in real-time. $G(s_{i}^{t})$ defines the Gaze potential for a shot $s_{i}^{t}$ at time $t$. The Fig. \ref{fig:Figure4} depicts the gaze potential dynamics within a single frame when considering the gaze of five distinct users. For gaze potential, a single human gaze path is sufficient for gaze potential when it's from a director. However, for improved gaze potential, multiple human eye gaze paths are needed.

\subsubsection{Editing cost}
We have enhanced the shot selection component to adapt to real-time settings, unlike the implementation in GAZED, which focused solely on offline processing. This process involves computing a cost matrix using gaze potential and incorporating penalty terms inspired by cinematic principles to avoid jump cuts, abrupt transitions, and other undesirable effects. We maintain the same penalty terms used in GAZED to construct the cost matrix. Moreover, the computation of the cost matrix is fast and can be performed in real-time.

To achieve shot selection in real-time, we leverage the ongoing construction of the cost matrix. At any given moment $t$, while the cost matrix is being built for a future time $t+f$ (with $f$ serving as a look-ahead duration in the cost matrix), we process the cost matrix information between time $t$ and $t+f$ to make shot selection decisions at time $t$. Now, let's delve into the details of the cinematically motivated penalty terms and the process of utilizing the cost matrix information from $t$ to $t+f$ for shot selection.

We use the following cinematically motivated penalty terms from GAZED \cite{Gazed20} - To minimize abrupt shot transitions, we introduce a transition cost ($T$). Smooth transitions require a low overlap between shot framing to avoid “jump cuts.” We also add a shot overlap cost ($O$) to prevent these abrupt jumps. The frequency of cuts impacts editing, where shot length influences audience perception. Longer shots evoke stillness, fitting emotional scenes, while shorter shots establish faster rhythms for energetic sequences. To regulate editing pace, we consider shot duration, leading to a cutting rhythm cost ($R$). These penalty terms contribute to the total cost of transitioning from one shot, denoted as $s^{t-1}$, to another shot, denoted as $s^{t}$. Both $s^{t-1}$ and $s^{t}$ belong to the set of available shots, denoted as $S$. The cumulative sum of these penalty costs determines the overall cost of transitioning between shots.

\begin{equation}
    E_{e}(s^{t-1}, s^{t}) = T(s^{t-1}, s^{t}) + O(s^{t-1}, s^{t}, \gamma) + R(s^{t-1}, s^{t}, \tau)
    \label{eqn:Equation5}
\end{equation}

We incorporate the defined penalty terms into the computation of the cost matrix ($C$). The cost matrix is constructed along the time dimension, where each cell represents the minimum cost required to reach that specific point. In building the cost matrix, we utilize recurrence relation - \ref{eqn:Equation9} that considers the information from the previous shot, the current shot's gaze potential, and other penalty terms. We can determine the optimal path and associated costs to navigate through the shots over time by evaluating this recurrence relation for each cell in the cost matrix.

\begin{equation}
    C(s_{j}^{t}, t) =
    \begin{cases}
        -ln(G(s_{j}^{t})), & t = 1 \\
        min_{i}(C(s_{i}^{t-1}, t-1) & otherwise \\
        \quad -ln(G(s_{j}^{t}))+E_{e}(s_{i}^{t-1}, s_{j}^{t})),
    \end{cases}
    \label{eqn:Equation9}
\end{equation}

To make the shot selection component work in real-time, we introduce Future and Continuity penalty terms. Computing GAZED on small windows of time frame lacks coherent shot selection between adjacent windows, leading to an unpleasant viewing experience. To ensure a coherent shot selection, we add penalty terms, addressing the issue of shot selection continuity.

The $Future$ penalty term $F$ encompasses the cost of choosing a specific shot $s_{k}^{t+f}$ at time $t+f$ when considering the entire path leading up to that point. We use the notation $s_{i}^{t}$ to represent a specific shot $s_{i}$ from the set of generated shots $S$ in a video frame at a given time $t$.
\begin{equation}
    F_{k} = C(s_{k}^{t+f}, t+f)
    \label{eqn:Equation10}
\end{equation}

Additionally, we introduce the method $Backtrack_{i}(z, t)$, which allows us to backtrack from a state $z$ at time $t$ to its preceding state $y$, which is behind $i$ time steps. This backtracking process enables us to identify the state $y$ that led to the current state $z$ during the forward pass in the cost matrix.

The $Continuity$ term in our methodology addresses the smooth transition between shots by penalizing the cost of switching from a previously selected shot, denoted as $p$, to a new shot $s$. To calculate this term, we perform a backtrack operation starting from a shot (state) $s_{k}^{t+f}$ in the future at time $t+f$. This backtrack operation allows us to trace back $f-1$ time steps and determine the cost of transitioning from the previously selected shot $p$ to the current shot $q$. Here, $q$ is obtained by applying the backtrack function.
\begin{equation}
    q = Backtrack_{f-1}(s_{k}, t+f)
    \label{eqn:Equation11}
\end{equation}

Considering the continuity term in our cost function (Equation - \ref{eqn:Equation12}), we ensure that the editing process maintains a seamless flow and avoids jarring transitions between shots. This term allows us to evaluate the transition cost between shots, considering the previously selected shot and the desired shot at a future time.
\begin{equation}
    Continuity_{k} = C(p, t) - ln(G(q)) + E_{e}(p, q)
    \label{eqn:Equation12}
\end{equation}

Now we devised the penalty terms to build a cost matrix. Rather than relying on a total cost matrix as in GAZED, We designed a real-time shot selection process that incorporates a small look-ahead duration ($f$) to make informed decisions. We follow a set of steps at each frame to determine the most suitable shot. We consider two parameters minimum shot duration $l$ and shot timer $\theta$ (which is crucial for the rhythm penalty term). We experimented with look-ahead ($f$) values of $32, 64, 128$. For higher look-ahead values ($f \geq 128$), Real Time GAZED generates video edits more like GAZED. Even with minimal look-ahead ($f=32$), Real Time GAZED performs comparably equal to GAZED.

Our real-time shot selection process ensures that each shot satisfies the minimum duration requirement $l$ to maintain coherence and avoid abrupt transitions. Additionally, we employ a shot timer $\theta$ to track the selected shots' duration. This timer is essential for accurately assessing the rhythm penalty, enabling us to maintain consistent pacing and flow throughout the video.

\begin{figure}
    \centering
    \includegraphics[width=0.75\linewidth]{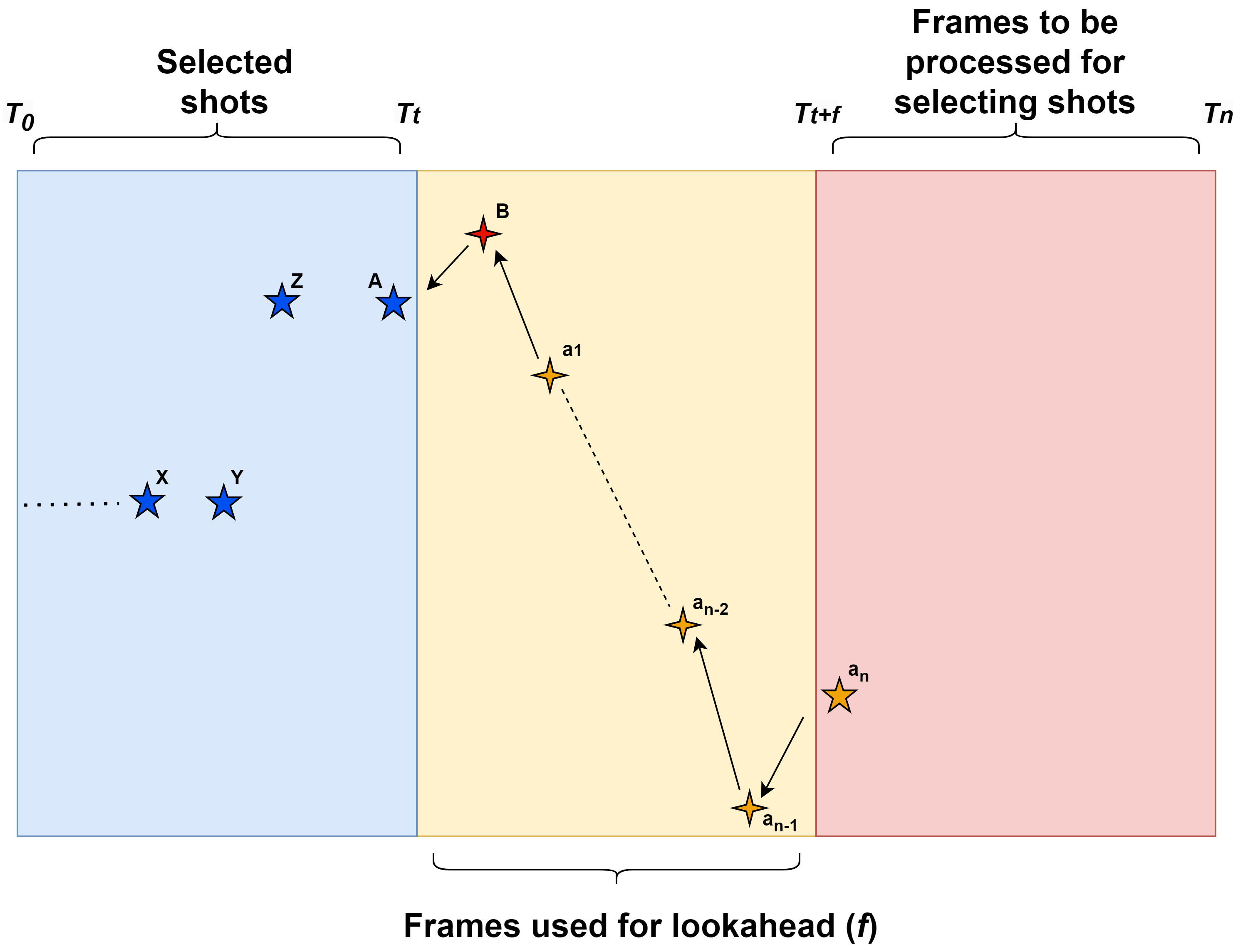}
    \caption{The figure provides a visual representation of how the cost matrix operates within the shot selection component of the Real Time GAZED pipeline. The blue region highlights the frames that had already been processed by the Real Time GAZED algorithm for shot selection. As indicated by the labels $X, Y, Z, \& A$ are several shots that have been previously selected within the timeframe from the start $T_{0}$ up to the current time $T_{t}$. The yellow region denotes the frames used for look ahead, providing a glimpse into the frames that are considered for future shot selection. The labels $a_{1}, a_{2}, ..., a_{n-1}$ represent the intermediary shots that are assessed by backtracking before the final shot $B$ is chosen at a time $T_{t+1}$. Finally, the red region represents the frames that are yet to be processed by the Real Time GAZED algorithm. Horizontal dimension denotes time $T$ and vertical dimension corresponds to the potential number of shots generated for each frame.}
    \label{fig:Figure5}
\end{figure}

\begin{enumerate}
    \item If the shot timer $\theta \leq l$ we adhere to a strict constraint. In this case, we select the previous shot as the current shot and increment the shot timer $\theta = \theta + 1$ to keep track of the elapsed time for the shot. In simpler terms, if the time allotted for a shot is within the predefined limit, we maintain continuity by keeping the same shot as the previous one.
    \item Else if the shot timer $\theta \geq l$, we proceed with the shot selection process by minimizing objective function (Equation - \ref{eqn:Equation13}). This objective function helps us identify the most suitable shot to select. Additionally, we reset the shot timer $\theta$ if the selected shot $s$ differs from the previously selected shot, ensuring a fresh start for the shot timer $\theta$. However, if the selected shot $s$ is the same as the previous shot, we increment the shot timer $\theta$ to continue the sequence, which gets penalized by rhythm cost $R$
    \begin{equation}
        min_{k}(F_{k} + \alpha*Continuity_{k})
        \label{eqn:Equation13}
    \end{equation}
    The objective function (Equation - \ref{eqn:Equation13}) is formulated as a combination of penalty terms that consider future shots and shot continuity. However, a tuning parameter, denoted as $\alpha$, is required to achieve the desired optimization. The term $F_{k}$ represents the cumulative cost of previous shots, which is calculated using a recurrence relation (Equation - \ref{eqn:Equation9}). Without careful consideration, the objective function may prioritize minimizing $F_{k}$ alone. When $\alpha$ is set to a higher value, the objective function is more likely to emphasize the continuity penalty term ($Continuity_{k}$).
    
    To strike a balance between the cumulative cost of future shots ($F_{k}$) and shot continuity ($Continuity_{k}$), it is advisable to choose $\alpha$ in proportion to the look-ahead duration ($f$). By adjusting $\alpha$ proportionally to $f$, We ensure that the optimization process considers the importance of the cumulative cost of previous shots and maintains continuity in a more balanced manner. In our experiments, we vary $\alpha$ within the range of $5$ to $10$. The Fig. \ref{fig:Figure5} illustrates the step-by-step shot selection process within the cost matrix in the shot selection component. 
\end{enumerate}

\begingroup
\setlength{\tabcolsep}{1.4pt}
\begin{figure*}
    \centering
    \begin{tabular}{c c c}
    \includegraphics[width=0.29\linewidth]{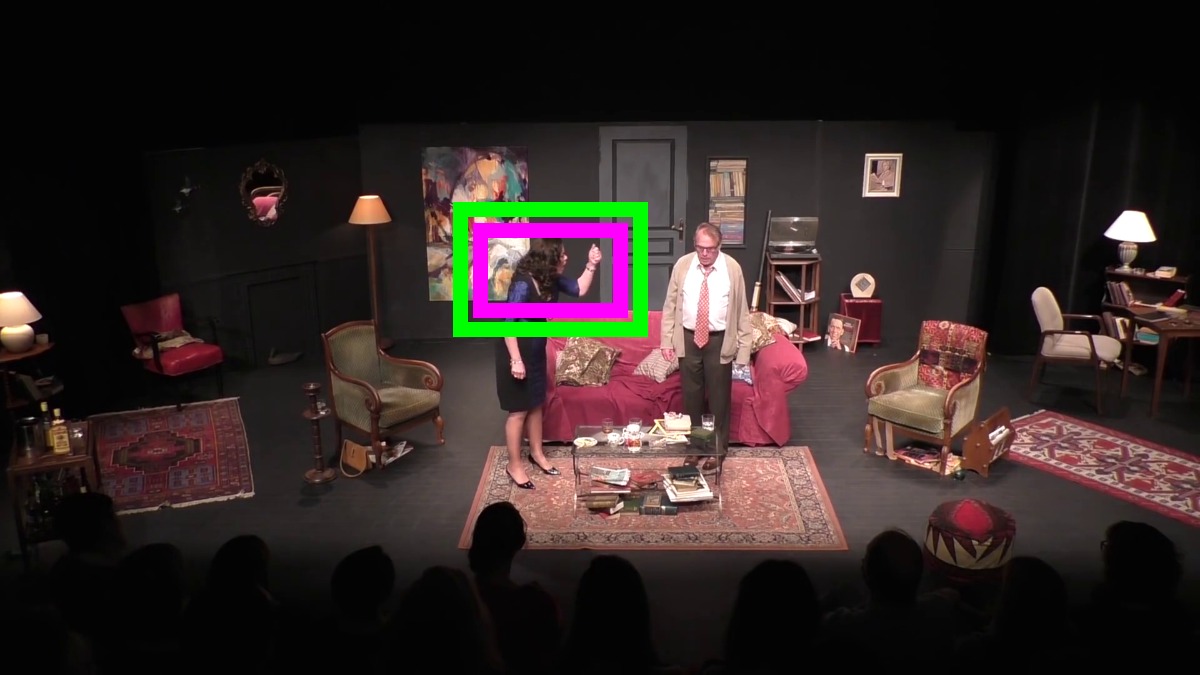} &
    \includegraphics[width=0.29\linewidth]{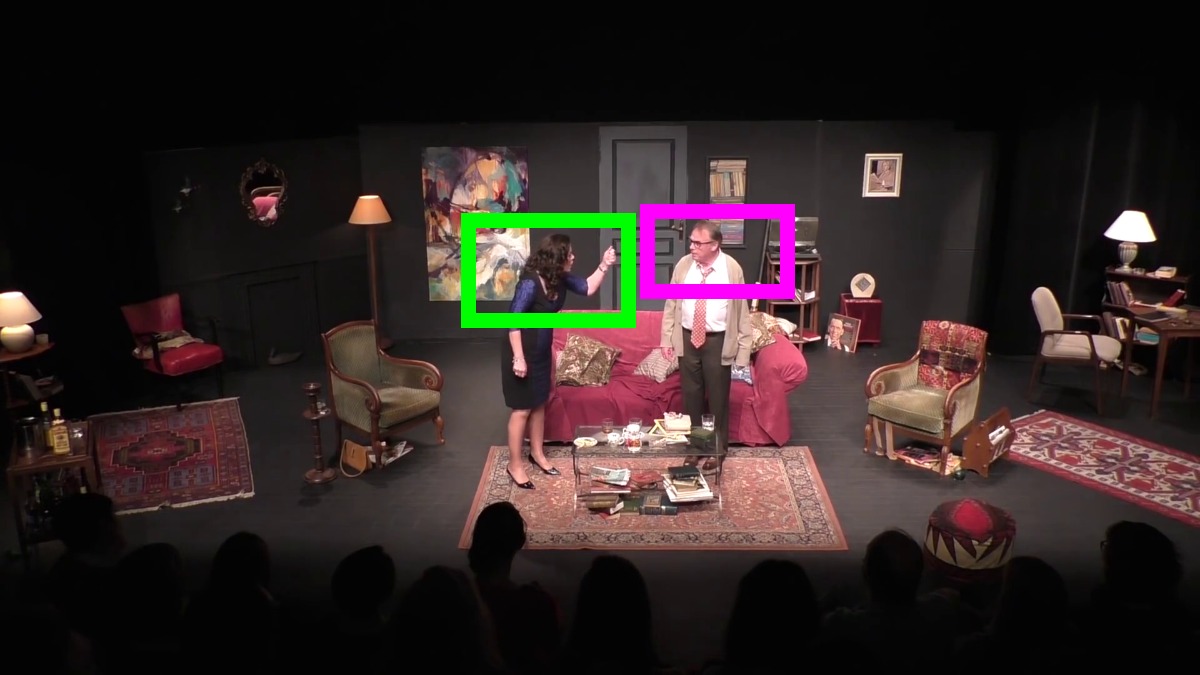} &
    \includegraphics[width=0.29\linewidth]{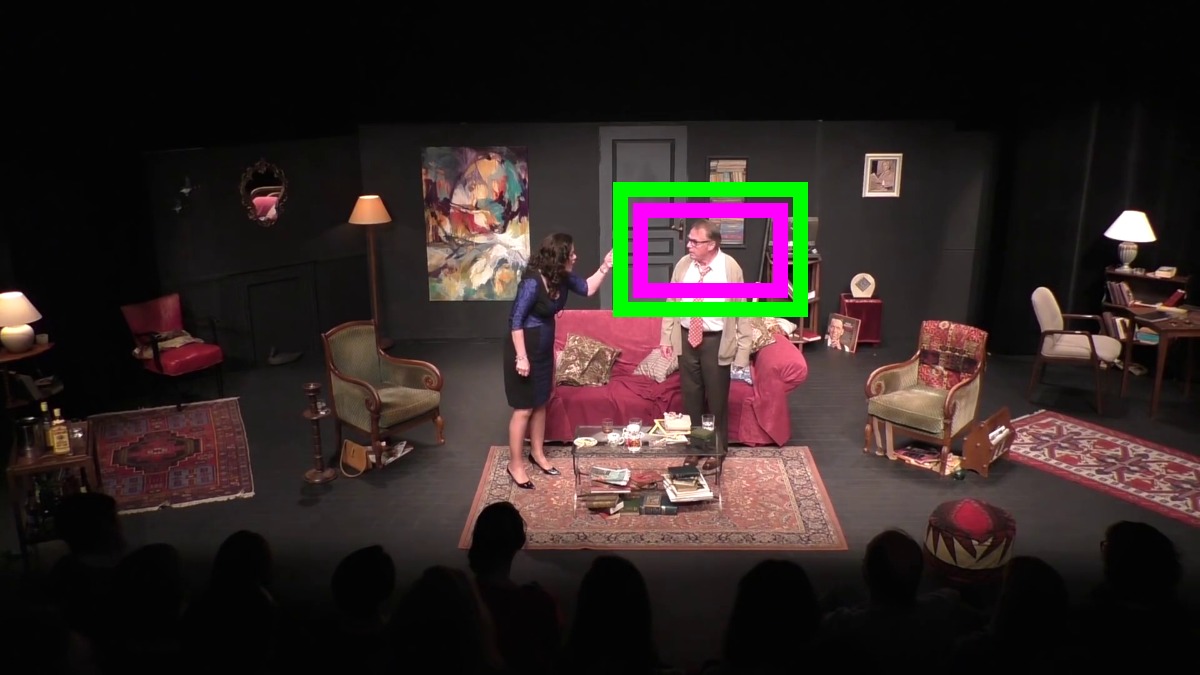}
    \end{tabular}

    \begin{tabular}{c c c}
    \includegraphics[width=0.29\linewidth]{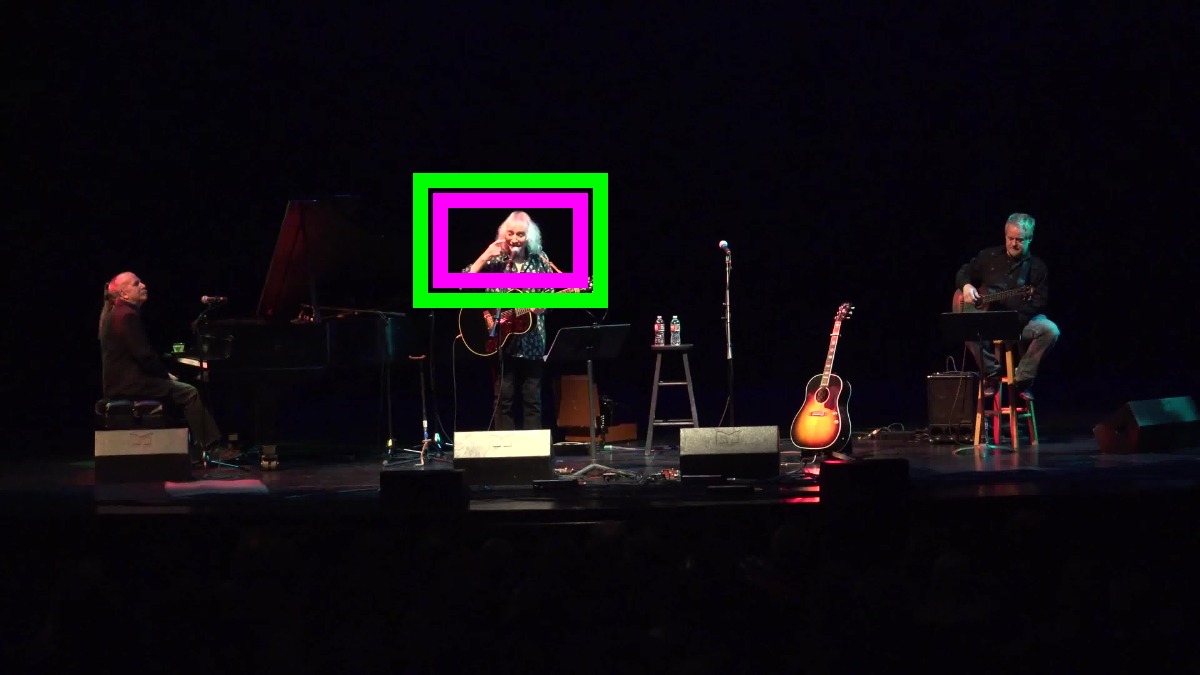} &
    \includegraphics[width=0.29\linewidth]{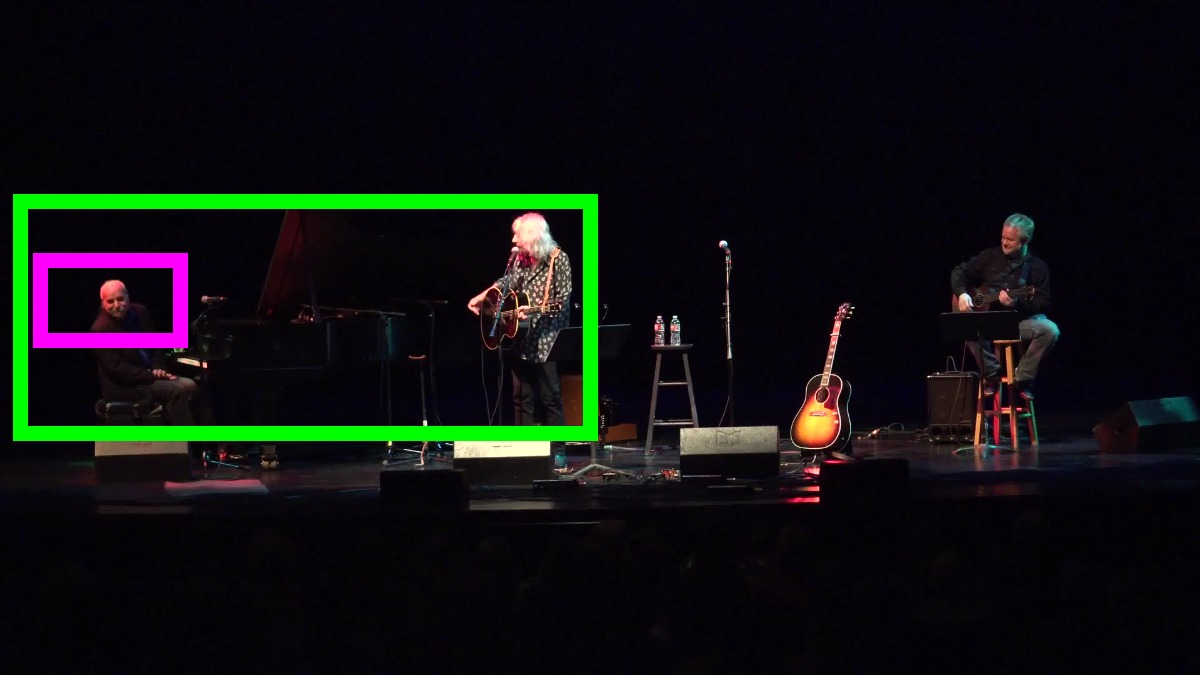} &
    \includegraphics[width=0.29\linewidth]{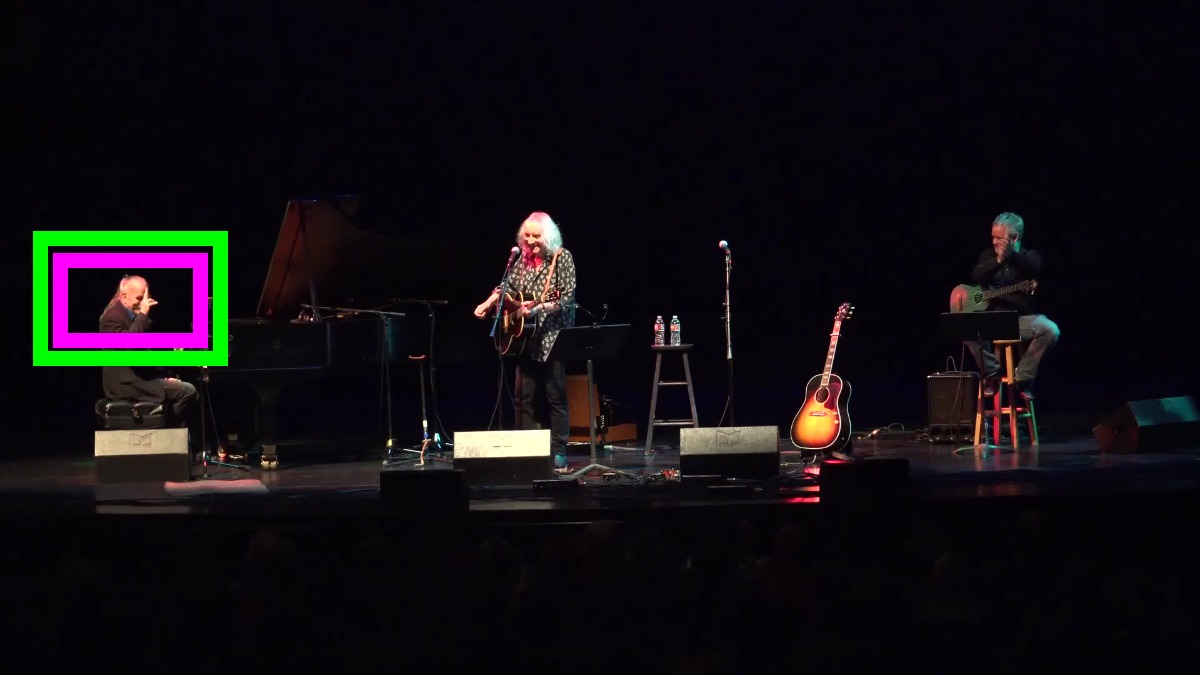}
    \end{tabular}
    
    \caption{The figure showcases a visual comparison of shot selections made by Real Time GAZED (highlighted in green) and GAZED (highlighted in pink) for two different videos. Shot selections made by Real Time GAZED exhibit a catching up behavior with GAZED, given that it operates in real time. While there may be intermediary differences in shot selection, as depicted in the middle column, Real Time GAZED dynamically adjusts its selection to minimize the overall cost and align with the shot chosen by GAZED.}
    \label{fig:Figure6}
\end{figure*}
\endgroup

\section{Comparison Baselines}
To evaluate the effectiveness of Real Time GAZED, we compare it against four video editing baselines: Wide, Greedy Gaze, Speaker-based, and GAZED itself. We set the minimum shot duration parameter ($l$) to 1.5 seconds to ensure a fair comparison.

\subsection{Wide}
The Wide baseline approach is inspired by the concept of video retargeting. It selects the widest shot possible, encompassing all performers on the stage.

\subsection{Greedy Gaze}
The Greedy Gaze \cite{Gazed20} editing algorithm greedily selects the shot with the highest gaze potential at each time instant $t$. However, since this approach solely relies on gaze information without considering cinematic editing principles, it may result in frequent shot switches that could hinder the understanding of the scene and degrade the overall viewing experience. To mitigate this issue, we enforce a minimum shot duration of 1.5 seconds (specified by parameter $l$).

\subsection{Speaker-based}

Speaker cues enhance dialog-driven scene editing \cite{ImprovingMeetingCapture} \cite{ComputationalVideoEditing}. Our Speaker-based baseline (Sp) selects the optimal shot for the speaker from rushes. Speaker data was manually annotated. Simultaneous speakers result in a combined shot. The chosen shot persists until a speaker change, with a minimum duration ($l$) to prevent quick transitions. Following an ablative study, our approach involves selecting a wide shot when a period of silence exceeding 10 seconds is detected.

\subsection{GAZED}
We also compare against the original GAZED framework, which serves as a baseline for our real-time version. It allows us to assess the improvements and performance of Real Time GAZED compared to the offline GAZED approach. In the Fig. \ref{fig:Figure6} provided, we can compare the shots selected by GAZED and Real Time GAZED approaches

\begin{figure*}
    \centering
    \begin{tabular}{c c c c}
        \includegraphics[width=0.2\linewidth]{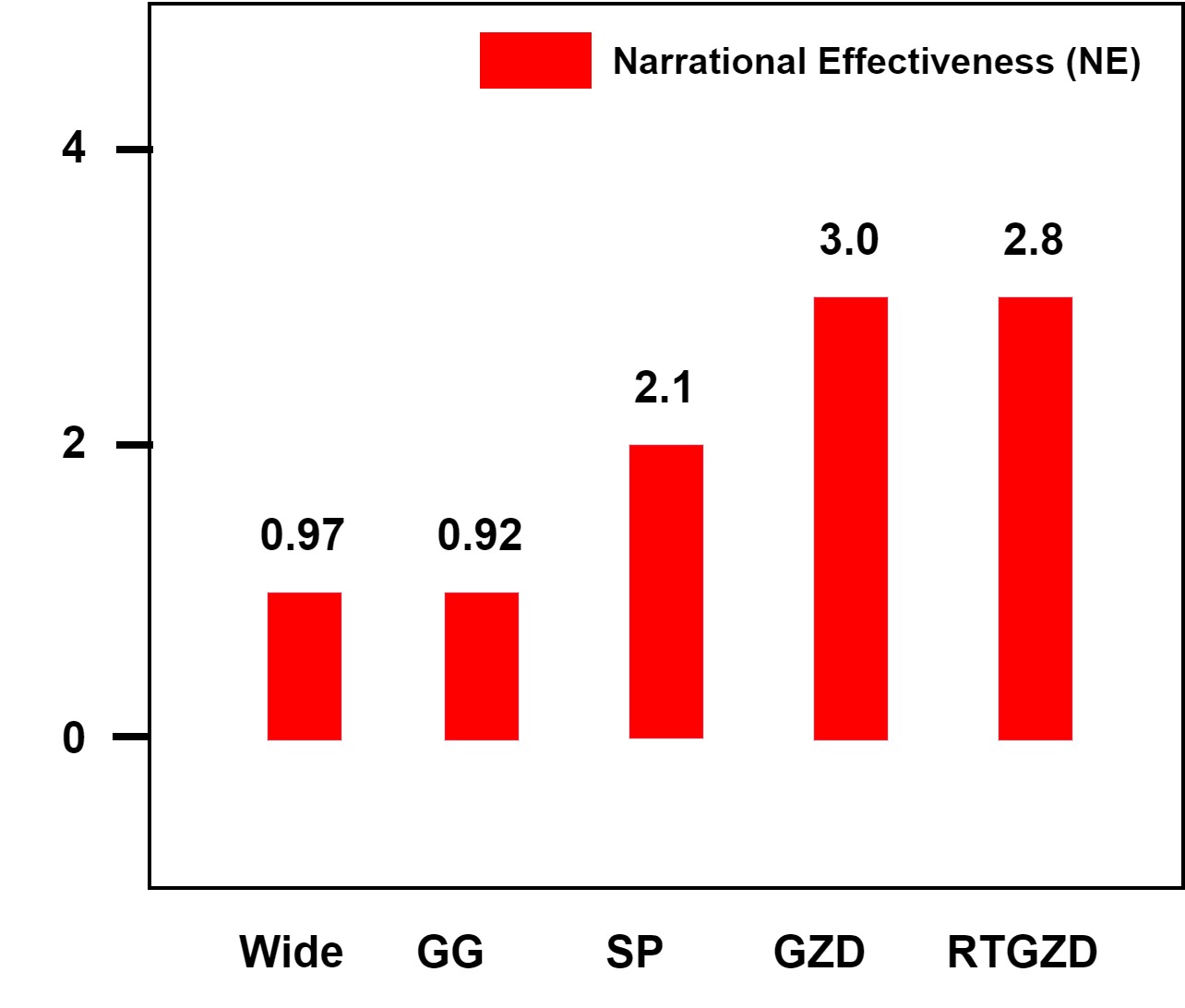} & \includegraphics[width=0.2\linewidth]{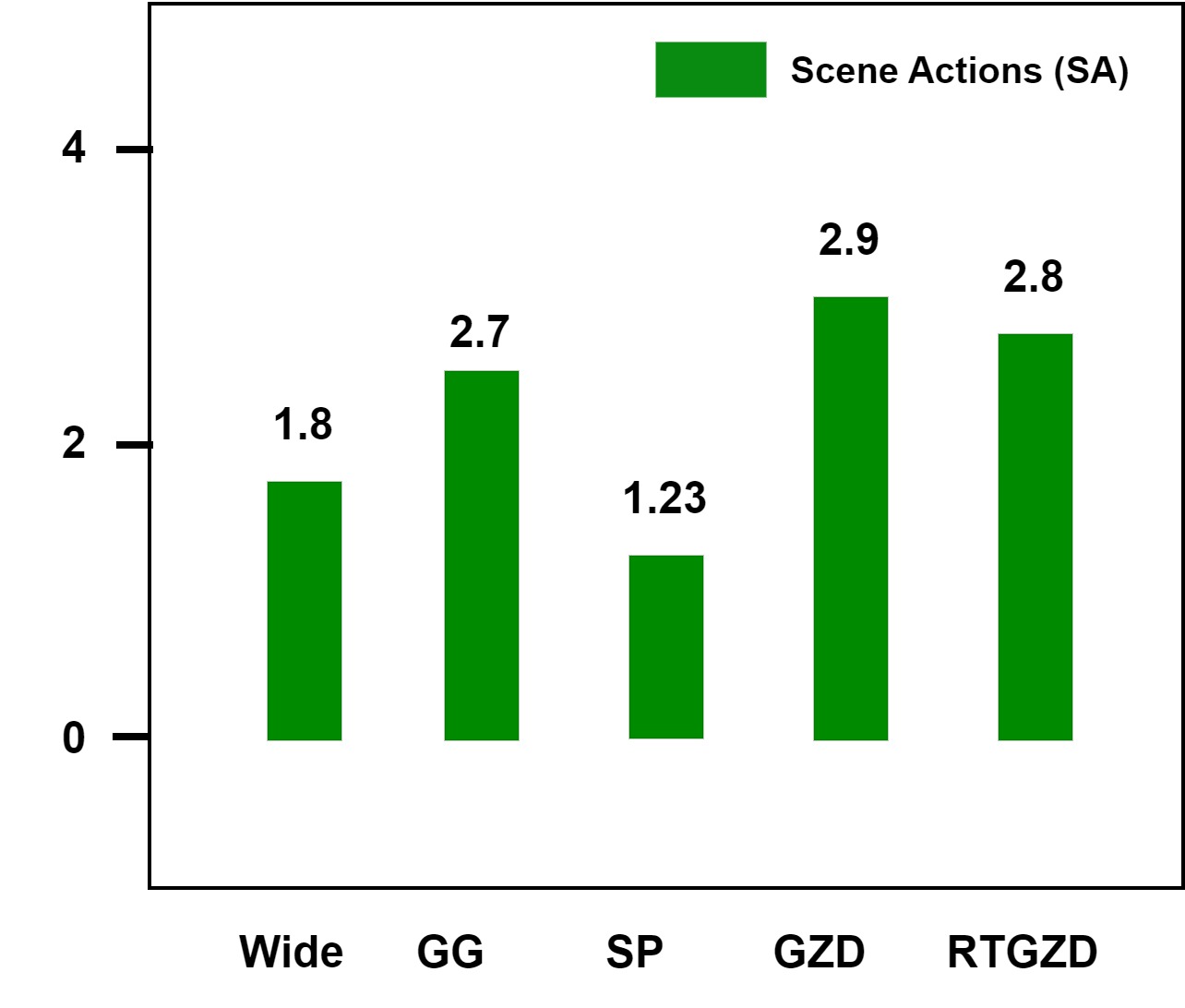} & \includegraphics[width=0.2\linewidth]{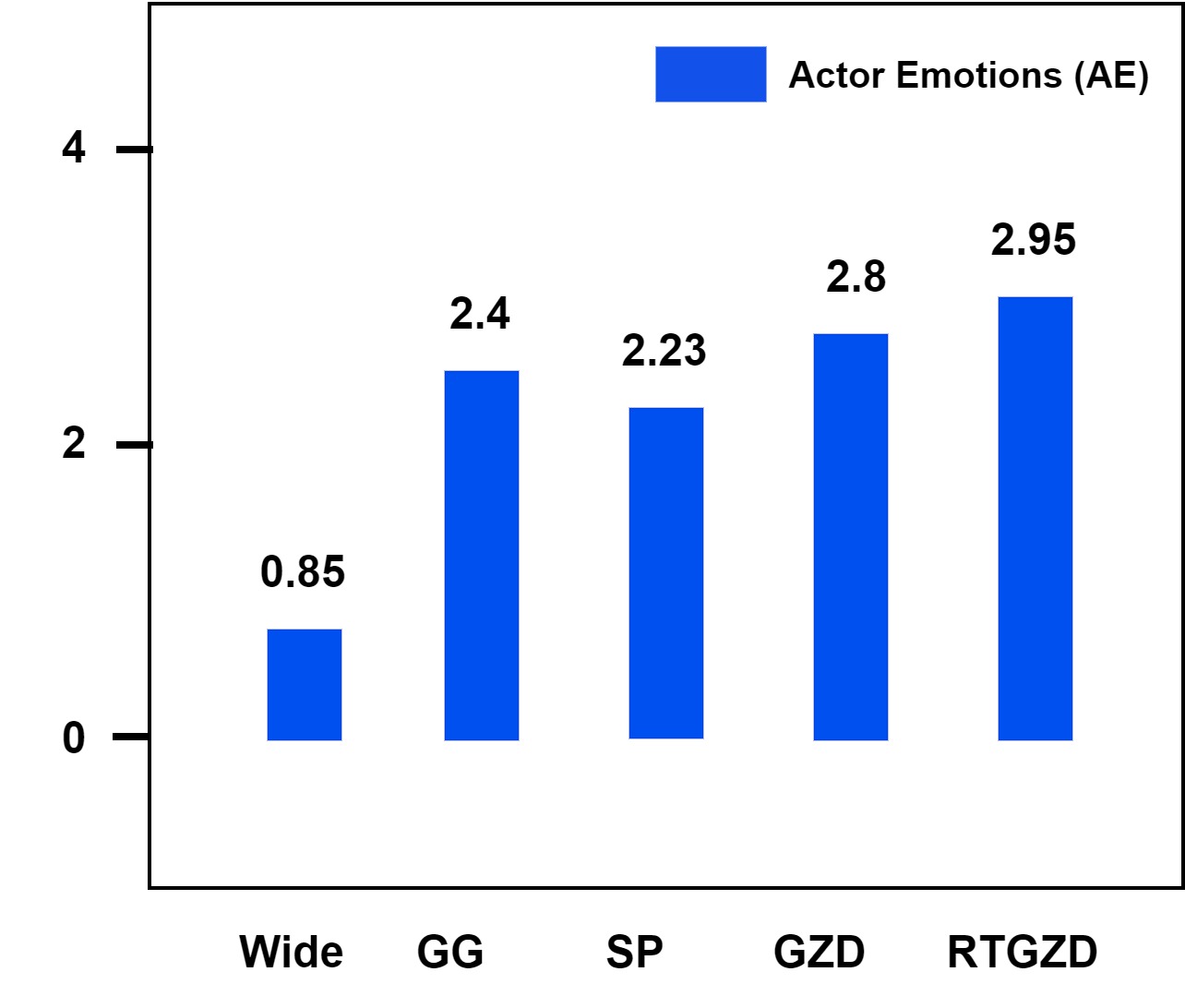} & \includegraphics[width=0.2\linewidth]{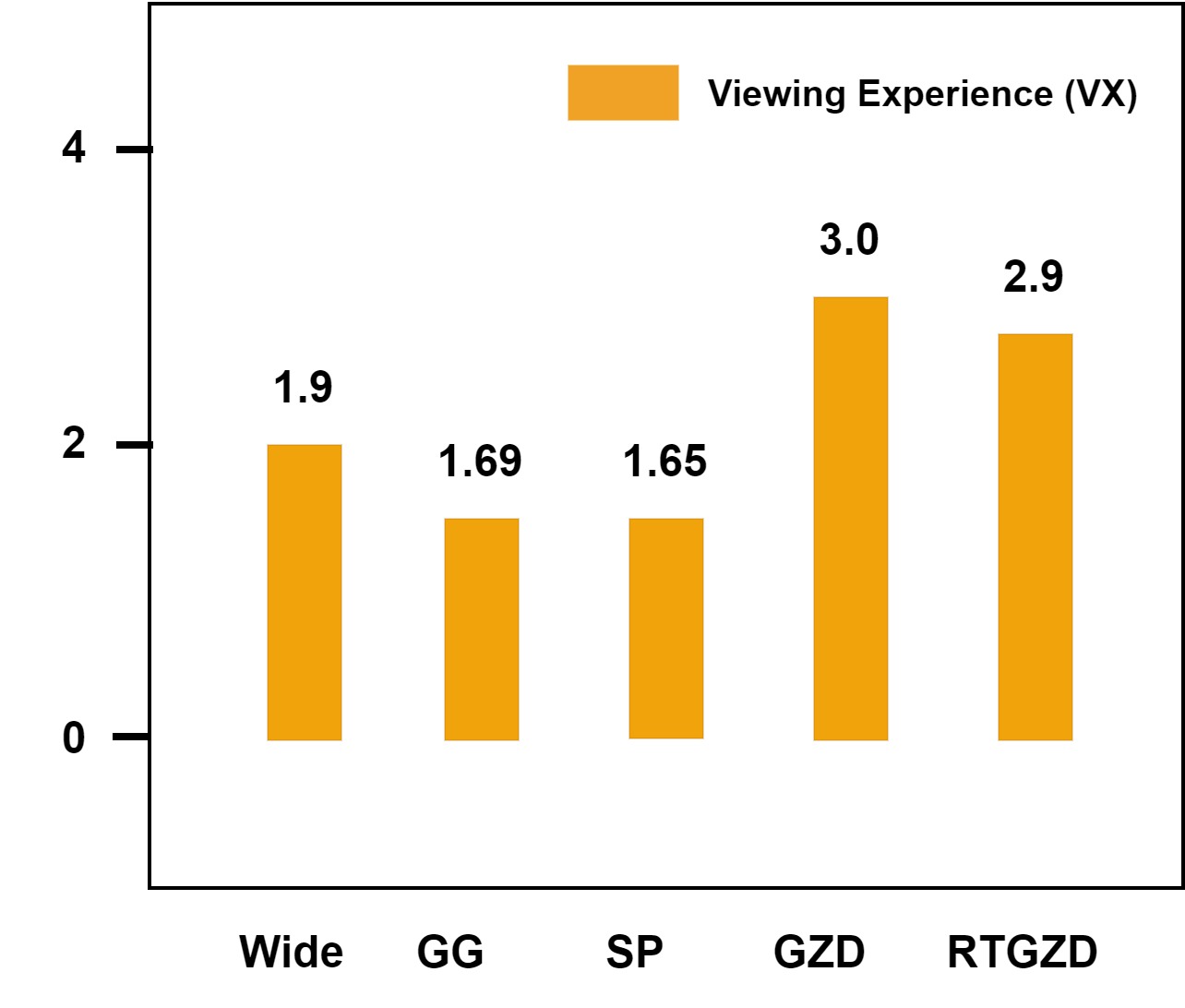}
    \end{tabular}
    \caption{Each bar in the histogram denotes the minimum and maximum user rating of narrational effectiveness (NE), scene actions (SA), actor emotions (AE), and viewing experience (VX) for each baseline Wide, Greedy gaze (GG), Speaker based (Sp), GAZED (GZD) and Real Time GAZED}
    \label{fig:Figure7}
\end{figure*}

\section{User Study}
To assess the video editing capabilities of GAZED compared to the baselines mentioned above, we conducted a user study involving 12 participants and 4 video recordings. Different editing strategies, including Wide, Greedy Gaze, Speaker-based, GAZED, and Real Time GAZED, were applied to generate edited versions of these videos. During the study, participants first watched the original video, followed by the randomly presented edited versions. We designed the study in a way that each participant viewed the original and edited versions of two-stage recordings, resulting in a total of 4 (different videos) × 2 (user ratings/video) × 5 (editing strategies) combinations. Participants were unaware of the specific editing strategy for each version they watched. After viewing each edited version, they were asked to compare it to the original and rate it on a scale of -5 to 5 for various attributes. The attributes of interest included:

\begin{enumerate}
    \item Narrational Effectiveness (\textbf{NE}): How effectively did the edited video convey the original narrative?
    \item Scene Actions (\textbf{SA}): How well did the edited video capture actor movements and actions?
    \item Actor Emotions (\textbf{AE}): How well did the edited video capture the actor's emotions?
    \item Viewing Experience (\textbf{VX}): How would you rate the edited
video for aesthetic quality?
\end{enumerate}

Prior to the study, participants were provided with information about the specific attributes and cinematic video editing conventions. They were then asked to rate each attribute using a scale relative to a reference score of '0' assigned to the original video. A positive score indicated that the edited version performed better than the original in terms of the specific attribute, while a negative score indicated that the edited version performed worse. The ratings provided by the participants were collected, and the mean scores for each attribute and editing strategy were calculated across all videos. The Fig. \ref{fig:Figure7} presents user ratings for all baselines, categorized according to each attribute.

\subsection{Narrational Effectiveness (NE)}
The Greedy Gaze (GG), Speaker-based (Sp), GAZED, and Real Time GAZED strategies, which prioritize actors and actions based on speech or gaze cues, outperform the Wide baseline in terms of their ability to capture the essence of the scene. The Wide approach often results in inefficient framing of the scene characters.

\subsection{Scene Actions (SA)}
The Wide and Speaker-based (Sp) baselines demonstrate similar performance in this aspect. These findings suggest that relying solely on speaker cues may be less effective in capturing focal events during stage performances. For instance, if one performer verbally introduces other co-performers to the audience, the Sp baseline may still prioritize the introducer instead of the introducee. In such cases, eye gaze turns out to be more accurate in capturing the events and actors of interest compared to speech. The Greedy Gaze (GG) strategy, which dynamically captures events of maximum interest at each time instant, effectively conveys scene actions and performs well.

\subsection{Actor Emotions (AE)}
The GG, Sp, GAZED, and Real Time GAZED techniques yield comparable performance. These methods effectively capture the speaker or leading actor in the scene through close-up shots, allowing for the clear conveyance of facial expressions and emotions to viewers.

\subsection{Viewing Experience (VX)}
As expected, GAZED and Real Time GAZED performed exceptionally well, receiving the highest scores for viewing experience among the five methods tested. The superiority of the Wide baseline over Greedy Gaze (GG) and Speaker-based (Sp) can be attributed to the fact that the Wide strategy ensures the entire scene context is always visible to the viewer. On the other hand, GG and Sp frequently cut between shots, focusing on perceived actions of interest, which can disrupt the viewing experience.

In our user study, a baseline video edit is rated negatively (below 0) if it fares poorly compared to the original (unedited) video. As our baselines aim to incorporate gaze or active speaker information, these edits were preferred over the original video, preventing any edit from receiving a score below 0, even if the rating fell in the range of -5 to 5. For quantitative analysis, we contrast Real Time GAZED's shot selections with GAZED as the benchmark. With minimal look-ahead ($f=32$), there's an $85\%$ average match on selected shots between the two methods. Increasing the look-ahead ($f=128$) raises the average percentage of selected shots match to $98\%$.

\section{Summary}
This study introduces Real Time GAZED, a modified version of the GAZED framework designed for real-time editing of stage performance videos. Real Time GAZED incorporates cinematic editing principles such as avoiding abrupt transitions, eliminating quick shots, and controlling the rhythm of shot changes by optimizing an energy minimization function with a small look ahead. The user opinions collected from a psychophysical study confirm the effectiveness of Real Time GAZED in producing visually pleasing and engaging edited videos. It provides competitive performance when compared against the original GAZED framework and outperforms the studied baselines. 

Real Time GAZED's real-time processing capability transforms editing into a dynamic task rather than a post-processing one. However, there are challenges associated with the real-time collection of human gaze data. The main limitation of our experiments is that it utilizes pre-recorded human gaze data, simulated in a causal manner. For estimating real-time human eye gaze, virtual reality headsets with built-in eye-tracking are suitable. Additionally, several companies provide wearable glasses with integrated eye-tracking sensors, like SMI ETG (Eye Tracking Glasses) and Tobii Pro Glasses. Certain smartphones have also begun incorporating basic eye-tracking functions, such as “Smart Scroll,” which employs your eye movements to scroll through content. In future work, we will explore integrating such real time gaze collection devices into the framework. Another future avenue would be to replace the gaze data with saliency prediction algorithms~\cite{jain2021vinet,jyoti2022salient}.


{\small
\bibliographystyle{ieee_fullname}
\bibliography{egbib}
}

\end{document}